\pdfoutput=1
\documentclass[letterpaper, 10 pt, conference]{ieeeconf}  

\usepackage{mathptmx} 
\usepackage{times} 
\usepackage{amsmath} 
\usepackage{amssymb}  
\usepackage{graphicx}
\usepackage{tabularx}
\usepackage{array}
\usepackage{diagbox}
\usepackage{xfrac}
\usepackage{multirow}
\usepackage{stackengine}
\usepackage{cite}
\usepackage{tikz}

\newcommand\copyrighttext{%
  \footnotesize \textcopyright 2021 IEEE. Personal use of this material is permitted. Permission from IEEE must be obtained for all other uses, in any current or future media, including reprinting/republishing this material for advertising or promotional purposes, creating new collective works, for resale or redistribution to servers or lists, or reuse of any copyrighted component of this work in other works.}
\newcommand\copyrightnotice{%
\begin{tikzpicture}[remember picture,overlay]
\node[anchor=south,yshift=10pt] at (current page.south) {\fbox{\parbox{\dimexpr\textwidth-\fboxsep-\fboxrule\relax}{\copyrighttext}}};
\end{tikzpicture}%
}

\newcommand{\fracentry}[2]{\addstackgap{\Large{\sfrac{#1}{#2}}}}

\DeclareMathAlphabet{\mathcal}{OMS}{cmsy}{m}{n}

\title{\LARGE \bf
Force-Sensing Tensegrity for Investigating Physical Human-Robot Interaction in Compliant Robotic Systems
}

\author{Andrew R. Barkan, Akhil Padmanabha, Sala R. Tiemann, Albert Lee, \\ Matthew P. Kanter, Yash S. Agarwal, and Alice M. Agogino} 

\begin{document}

\maketitle
\copyrightnotice
\thispagestyle{empty}
\pagestyle{empty}

\begin{abstract}

Advancements in the domain of physical human-robot interaction (pHRI) have tremendously improved the ability of humans and robots to communicate, collaborate, and coexist. In particular, compliant robotic systems offer many characteristics that can be leveraged towards enabling physical interactions that more efficiently and intuitively communicate intent, making compliant systems potentially useful in more physically demanding subsets of human-robot collaborative scenarios. Tensegrity robots are an example of compliant systems that are well-suited to physical interactions while still retaining useful rigid properties that make them practical for a variety of applications. In this paper, we present the design and preliminary testing of a 6-bar spherical tensegrity with force-sensing capabilities. Using this prototype, we demonstrate the ability of its force-sensor array to detect a variety of physical interaction types that might arise in a human context. We then train and test a series of classifiers using data from unique and representative interactions in order to demonstrate the feasibility of using this physical modality of sensing to reliably communicate goals and intents from a human operator in a human-robot collaborative setting.

\end{abstract}

\section{Introduction}
Physical interaction is an inevitable aspect of human-robot cooperation. In order to successfully integrate robots into a human environment, it is critical to consider the physical consequences of their existence and operation in such contexts. Physical interactions can even serve to facilitate cooperation by acting as a mode through which humans and robots can understand each other's states, behaviors, and intents. The physical characteristics of traditional rigid robotic systems often limit their functionality in scenarios involving contact with humans and within human environments. Instead, roboticists have investigated a broad range of \textit{compliant} robotic systems that are designed with intrinsic non-rigid characteristics that make them more applicable to contexts that involve physical interaction. An example of one class of compliant robotic systems is \textit{tensegrity} robots.
\par
\begin{figure}
    \centering
    \includegraphics[width=0.48\textwidth]{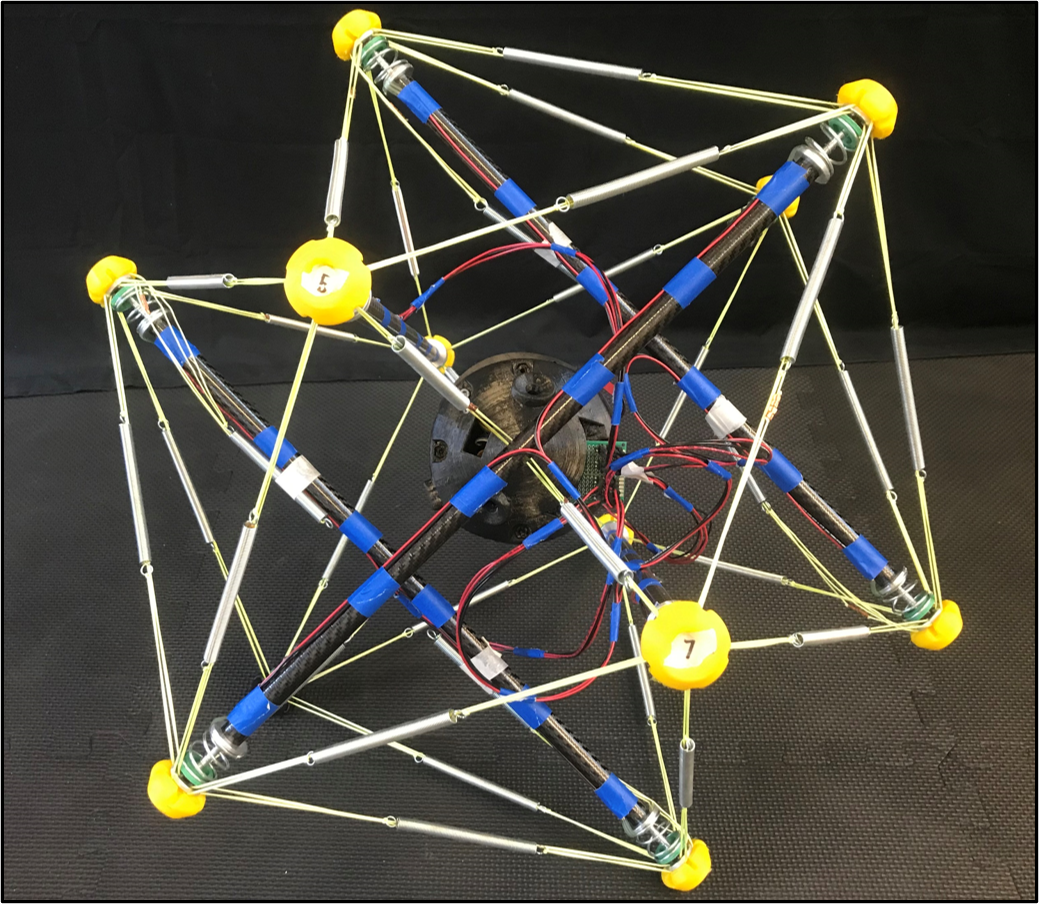}
    \caption{The prototype force-sensing tensegrity for investigating physical human-robot interaction with compliant robots.}
    \label{fig:robot}
\end{figure}

Tensegrity (tension-integrity) robots are robotic systems that leverage structures based on tensegrity principles \cite{skel2009}. Classically employed in structural engineering and art, tensegrity as applied in robotics has led to many useful realizations that feature properties such as structural compliance, robustness, and configurability \cite{sabel2015,fries2014,bruce2014,booth2020}. These unique properties found in tensegrity structures create the opportunity to develop systems that are highly tolerant to physical interactions. Furthermore, this tolerance can be potentially exploited in a way that benefits a system's capacity for physical human-robot interaction (pHRI). We envision a class of physically-tolerant, compliant robotic systems that can perceive and interpret physical interactions in ways that allow for intuitive human-robot interfaces and improvements in cooperative task efficiency.
Specifically, this study will feature the design of a compliant 6-bar spherical tensegrity instrumented with force-sensing capabilities. An image of the force-sensing 6-bar spherical tensegrity prototype is shown in Figure \ref{fig:robot}.
\par
\begin{figure}
    \centering
    \includegraphics[width=0.48\textwidth]{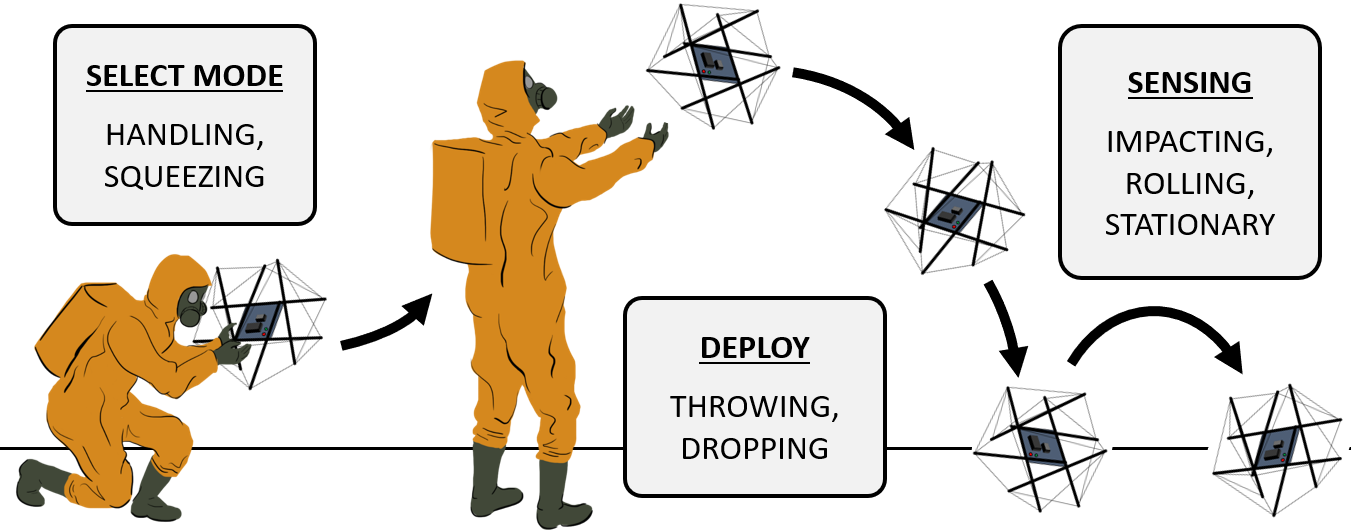}
    \caption{Illustration of a sequence of physical human-robot interactions involving three different stages of operation between a first responder and a compliant mobile tensegrity robot.}
    \label{fig:scenario}
\end{figure}

Here, we consider the illustrative example of first responders deploying a compliant tensegrity robot to safely and remotely assess a disaster scenario. In this scenario, non-technician operators are expected to interface with robots through a bulky HazMat suit while navigating a high-stress environment that can involve visual obstructions and auditory distractions. Instead of relying on non-intuitive remote controls or other complex user interfaces, compliant robots can infer an intended task through physical perception in real-time. The image shown in Figure \ref{fig:scenario} illustrates an imagined scenario in which first responders in HazMat suits employ a sequence of intuitive physical interactions to operate a compliant mobile tensegrity robot without the need for direct manual control or programming.

The paper is organized as follows. First, we will discuss the significance of force-sensing in the context of compliant robotic systems directed at improving approaches to pHRI. Next, the design of the first force-sensing tensegrity prototype will be presented. We will follow this by describing our approach to distinguishing physical interactions by comparing modern classification modeling techniques. Finally, we will examine the effectiveness of our classifiers using real data from the force-sensing tensegrity.

\section{Prior Work and Background}

In most applications of force sensing in robotics, the actuation platform is comprised of a rigid architecture that has been instrumented with sensing either preemptively in the design of its actuation mechanism as seen in series elastic actuators \cite{yu2015,calan2014}, or retroactively by instrumenting its end effector or other interfacing components, which are common in industrial settings \cite{lange2013}. In these implementations, there is an inherent need to transform the local measurements into a meaningful force reading in the context of the whole robot (e.g. through forward dynamics). The latter case requires roboticists to integrate sensing solutions into the components of a robot that interface with its environment, or in the case of pHRI, with humans. Perhaps the most easily accessible application of this type of approach to pHRI is the realization of tactile sensing in robotic systems \cite{argall2010,ciri2015}.
Recently, the growth of pHRI as a field of research has introduced force sensing as a means to facilitate more effective interaction with humans, and more generally, their environments \cite{sant2008,hadd2016,losey2018}. Having direct access to the dynamics of an interaction has enabled intelligent systems to reason about the consequences of these forces and then react appropriately, such as in human-robot collaborative tasks \cite{agra2014,bajc2018}.
\par
Research in pHRI has been disproportionately devoted to applications involving robots that fit into a set of common archetypes such as anthropomorphic arms. More recently, the advent of advanced soft robotic systems has enabled innovative studies focusing on interactions that take advantage of the characteristics of compliant mechanisms \cite{sanan2011,poly2017,maj2014}. However, one category of robotic systems that has yet to receive sufficient attention in pHRI is mobile robotic systems. This is primarily due to the sensitivity of rigid mobile robots to contact. Indeed, the paradigmatic approach to mobile robot control is to avoid contact at all costs.
\par
In efforts to design for intrinsic robustness in mobile robots, roboticists have drawn inspiration from tensegrity structures to develop mobile robots that are capable of being thrown from a helicopter or plane and still operate \cite{squish2019}. In the context of our discussion on pHRI, the development of such compliant mobile robotic systems has important implications for how we design accompanying human-robot interfaces. Equipped with the ability to physically interact with compliant mobile robots, we are given access to a whole new set of possible physical interaction mechanisms that can potentially expand the ability of human and robot to communicate and cooperate through contact. We see the development of a force-sensing tensegrity capable of differentiating basic physical interactions as the first step to understanding a new language of pHRI with compliant robots.

\section{System Design}



A tensegrity structure is a structure composed of a set of rigid elements that are held in static equilibrium via a set of connecting cable elements so that the whole structure is suspended in what is called a stable tensegrity configuration \cite{skel2009}.
Tensegrity structures exhibit properties that make them uniquely suited for applications involving physical interaction. The intrinsic robustness to external perturbation enables tensegrities to withstand contact and even violent impacts without sustaining critical damage \cite{squish2019}. At the same time, the rigid elements in the structure constitute a suitable scaffolding for instrumentation, which can be exploited to better sense how the structure responds to external physical stimulus. Furthermore, the connectivity of a class one tensegrity affords the structure the useful property of exclusively axial loading on rigid elements. The fact that rigid elements are subjected only to axial loading has important consequences for the design of a force-sensing tensegrity for investigating pHRI, which we will now discuss.
\par
In terms of structural properties, the lack of bending loads on rigid elements of the tensegrity allows for the use of much lighter components with lower density material. One of the main reasons tensegrity principles are employed in structural design is for their high ratio of strength to member density. In relation to pHRI, low weight is an attractive characteristic because it translates to low inertia, which contributes to the inherent manipulability and safety of a system whose express purpose is to be in physical contact with humans.
\par
Perhaps even more important than the structural advantages gained from exclusively axial loading are the benefits to force-sensing instrumentation. Because of the interconnectedness of a tensegrity, any force that acts to deform the structure will result in a detectable change in axial loading on some combination of the members. Thus, external physical perturbations to a tensegrity are detectable via instrumentation of its internal members \cite{sult2004}.

\subsection{Spherical 6-Bar Tensegrity}

The spherical 6-bar tensegrity presented in this study is a class one tensegrity consisting of 6 rigid elements and 24 cable elements arranged so that the convex hull formed by the \textit{nodes} (or connecting points between members) of the members form an icosahedron shape. Our implementation of this tensegrity topology consists of cylindrical carbon fiber tubes as rigid elements and assemblies of steel springs and string (connected in series) acting as cable elements. The elastic cable assemblies give the entire tensegrity structure a configurable degree of compliance, which has been selected to create a device that can deform in response to physical perturbations. A centrally located payload suspended via cable elements from the nodes of the structure houses and protects embedded electronics systems for sensor integration, data logging, and wireless communication. The fully assembled and integrated tensegrity has an approximate diameter of $0.56$m and a mass of $0.70$kg.
\par
The force-density method used in \cite{fries2014} is a common approach to determining equilibrium conditions for network structures. As demonstrated in their treatment of the force-density method, there is a linear relationship between internal member loading and external forces on the nodes of the tensegrity structure. This can be useful property for extracting information about external conditions on the structure through internal instrumentation. Depending on choice of connecting members, however, this linear mapping between internal loading and external forces generated for a given tensegrity configuration and set of node positions is not necessarily injective. As a result, our ability to deduce external forces in the static case is potentially limited, even with full access to axial load and member length measurements through the instrumentation of the structure. Even so, it has been demonstrated that a statics-based optimization approach can be effective for estimation of tensegrity states \cite{sabel2020}. Dynamic analyses, on the other hand, are forced to make restrictive assumptions surrounding external constraints in addition to assuming full knowledge of node positions \cite{skel2009}. In general, practical estimation of tensegrity states is a challenging problem due to a lack of constraints and limitations on sensing integration \cite{tur2009}.
\par
Instead of attempting to achieve full access to both position and force data of the entire structure, we propose aiming for partial knowledge of internal loads by instrumenting a subset of members. This will allow us to develop an empirical physical interaction model by leveraging machine learning in a classification-based approach. In this way, we will be able to exploit the detectability of external forces through the internal members without relying on a restrictive theoretical model of the structure to deduce interactions. The next section will describe how force sensor instrumentation was achieved to offer the most comprehensive detection of physical interactions in a minimal package.

\subsection{Force-Sensor Array}

There are numerous examples of force sensors being integrated into the structure of a tensegrity. The majority of approaches opt for instrumenting the cable elements of the structure due to ease of replacement and control considerations. However, there are drawbacks to using the cables to integrate force sensors. If the structure is deformed in such a way that one or more of the cable elements lose tension, the associated force readings on those affected cables will be rendered temporarily unavailable, which can potentially limit how well the system is able to detect physical interactions. Furthermore, the high flexibility and low inertia of cable elements necessitate the use of extremely lightweight sensors to avoid influencing force readings.

\begin{figure}
    \centering
    \includegraphics[width=0.48\textwidth]{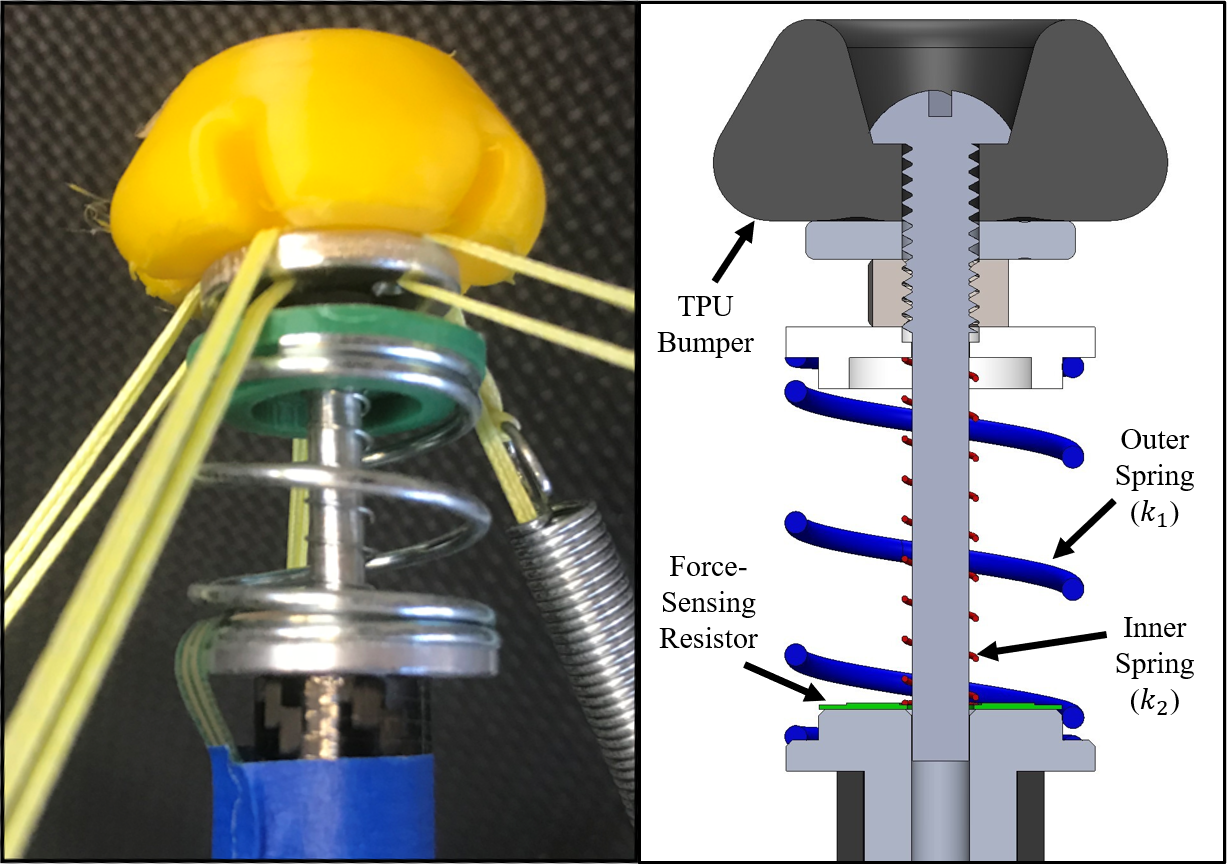}
    \caption{(left) One of twelve FSR-integrated nodes in the prototype force-sensing tensegrity. (right) Section view of node assembly highlighting force divider mechanism.}
    \label{fig:node}
\end{figure}

For our implementation of a force-sensing tensegrity, we instrument the rigid bars of the structure using a novel tensegrity node assembly shown in Figure \ref{fig:node}. The figure shows a cross-section of one node assembly that is integrated into each end of the 6 rigid bar elements of the structure. Notice that the node itself is not rigidly connected to the bar element but is instead floating by way of a linear rod passing into the hollow opening of the bar. The assembly is held together by the compressive forces from the equilibrium state of the tensegrity structure. Since the rigid elements of a stable tensegrity configuration are always held in compression, the force sensors are sensitive to changes in compressive load as a result of external forces to any member, including both cable elements and rigid elements, in the structure. The entire bar assembly is made resilient to axial impacts through the use of a 3D-printed TPU bumper on each node.
\par
In order to accommodate the use of low-profile and cost-effective Interlink FSR-404 force-sensing resistors (FSRs) while still satisfying force sensing requirements, it was necessary to design a mechanism by which we could extend the force-sensing range. We created a mechanical realization of a voltage divider by leveraging two compression springs in parallel. The stiffer of the two springs resists higher loads while the more compliant spring transmits force to the FSR. Since the springs are arranged in parallel, the displacement of the mechanism can be used to convert the actual force $F$ to an accurate reading $F_{meas}$ using the following equation.

\vspace{-0.1cm}
\begin{equation}\label{eqn:force}
F = \left(\frac{k_{1} + k_{2}}{k_{2}} \right)F_{meas}
\end{equation}

We could then configure the measured force range of $F_{meas}$ by selecting an appropriate set of springs in this assembly ($k_1=3.78$N/mm, $k_2=0.37$N/mm). Because the node assemblies are only held together by compression, it is straightforward to replace and configure parts to adjust the sensor integration. In order to account for the added dynamics of the node assemblies, we instrument the rigid bars on both ends resulting in a total of 12 force sensor inputs. Next, we will present our approach to physical interaction inference using classification modeling.

\section{Classification Modeling}

In order to verify the feasibility of our proposed sensing implementation, we employ contemporary multiclass classification modeling approaches to experimental data across a set of physical human-robot interactions that are unique to contexts involving compliant robotic systems. We hope to demonstrate through supervised learning that the system provides sufficient information to reliably distinguish between physical interaction types. We also explore the implications of signal time dependence on classifier inputs.

\subsection{Problem Definition}

To test our hypothesis, it is necessary to first establish a set of representative interaction types that explore the potential for novel pHRI with compliant mobile robotic systems. As an illustration, we can consider a scenario in which a human operator is attempting to communicate with a compliant mobile robot that is capable of sensing contact forces. One sequence of events could involve the human operator picking it up, squeezing it, and tossing it in order to communicate a desired intent as in the hypothesized disaster response scenario shown in Figure \ref{fig:scenario}. Each of these individual actions on the part of the human operator constitutes an interaction that is unique and detectable by compliant robots.
\par
We define the following four distinct classes of physical interactions to be classified using inputs from the force-sensing array in our prototype tensegrity: (1) {\it null}, (2) {\it drop}, (3) {\it squeeze}, and (4) {\it handle}. In a later section, we will describe the interactions in more detail along with experiment protocols for gathering train and test data sets on each of the aforementioned classes through real-life demonstrations with human operators and the prototype tensegrity.
\par
Let $(\mathcal{X}_1, y_1), (\mathcal{X}_2, y_2), ... (\mathcal{X}_n, y_n)$ be a set of $n$ physical interaction observations where each instance is comprised of a set of input features $\mathcal{X}_i \in \mathcal{D}$, where $\mathcal{D}$ is the space of features, and corresponding class label $y_i$. The goal is to learn a function $y_i=f(\mathcal{X}_i)$ that maps a set of input features $\mathcal{X}_i$ to its correct label $y_i$. There is a broad range of possible approaches to learning this mapping using contemporary classification algorithms. In the next section, we discuss choice of algorithm and time series classification factors.

\subsection{Algorithms and Feature Engineering}

We compare a combination of different processing pipelines and choices of algorithm in order to analyze the feasibility of solving the previously discussed multiclass classification problem and the robustness of our solution. Since the intent of pHRI in this context is to facilitate real-time intent inference from the perspective of the robot, we will also explore the influence of time on our model's ability to classify interaction types reliably.
\par
Each physical interaction constitutes a set of signals that represent changes in force over time across each of the 12 nodes of the tensegrity structure. In order to investigate temporal effects, we process the signals from each interaction by sectioning them into individual observations based on a certain window size so that a newly sectioned observation only sees a fraction of the entire interaction. By varying the size of this window, we can better understand the effect of time on the amount of information contained in the force sensor readings from an observation. As a second step, we choose between two different feature engineering approaches: (1) raw time series features and (2) statistical and temporally abstracted features.
\par
In the first case, which we will refer to as the ``raw" feature processing approach to indicate raw time series, we treat each instantaneous force value in a time series as a feature input across all force sensors. The second case, which we call the ``abstract" feature processing approach to indicate a second processing step, involves calculating a set of values that are intended to capture critical information from force data while abstracting away time dependence. The feature values are calculated as follows:
\begin{itemize}
    \item {\it total impulse ($J$)} : Discrete time integral of force over a given time window.
    \begin{equation}
        J=\sum_{t}\frac{\Delta t}{2}(F(t_{i-1}) + F(t_{i}))
    \end{equation}
    \item {\it maximum yank ($Y_{max}$)} : Maximum discrete time derivative of force wrt time over a given time window.
    \begin{equation}
        Y_{max}=\max_{t}\frac{(F(t_{i+1}) - F(t_{i}))}{\Delta t}
    \end{equation}
    \item {\it maximum force ($F_{max}$)} : Maximum force over a given time window.
    \begin{equation}
        F_{max}=\max_{t}F(t)
    \end{equation}
\end{itemize}
Each feature is intended to capture representative aspects of the force signals recorded from each kind of physical interaction to aid the classifier in distinguishing between classes. For example, we expect to observe higher maximum yank in {\it drop} interactions than other types due to the force of impact. The above set of features are calculated for each force sensor resulting in $36$ total features, which remains consistent across varying window sizes. Next, we will discuss choice of algorithm.
\par
We employ two common algorithms for training a multiclass classifier based on our previously described framework: k-Nearest Neighbors (KNN) and Random Forest (RF). The k-Nearest Neighbors algorithm is a non-parametric approach to training classifiers that is often an initial choice as a baseline algorithm for assessing the performance of classification frameworks \cite{aly2005}. We also apply the Random Forest algorithm, which is an ensemble learning method based on decision trees \cite{liaw2002}. The RF algorithm has been shown to outperform many alternative multiclass algorithms in problems that involve large data sets with high dimensionality and will be an excellent test of classification potential in this context.


\section{Experiments}

The experimental setup is designed to gather data from the aforementioned set of physical interactions between our force-sensing tensegrity and a human operator. The intent of these experiments was to accumulate testing and training data sets to verify the performance of our classification framework, and thereby justify the feasibility of the proposed approach to pHRI with compliant mobile robots.
\par
For each of the four types of physical interactions, a human operator performs a sequence of predetermined actions involving physical contact with the prototype force-sensing tensegrity. In the {\it drop} interaction, the human operator holds the device at a height of $1$m and releases it, allowing it to impact the ground. The {\it squeeze} interaction involves having the human operator compress the tensegrity between two rigid surfaces in a given orientation. The {\it handle} interaction represents data from a human operator holding the device and turning it one full rotation in their hands. As the name suggests, {\it null} interactions involve the device sitting on a surface without any human contact.
\par
\begin{figure}
    \centering
    \includegraphics[width=0.48\textwidth]{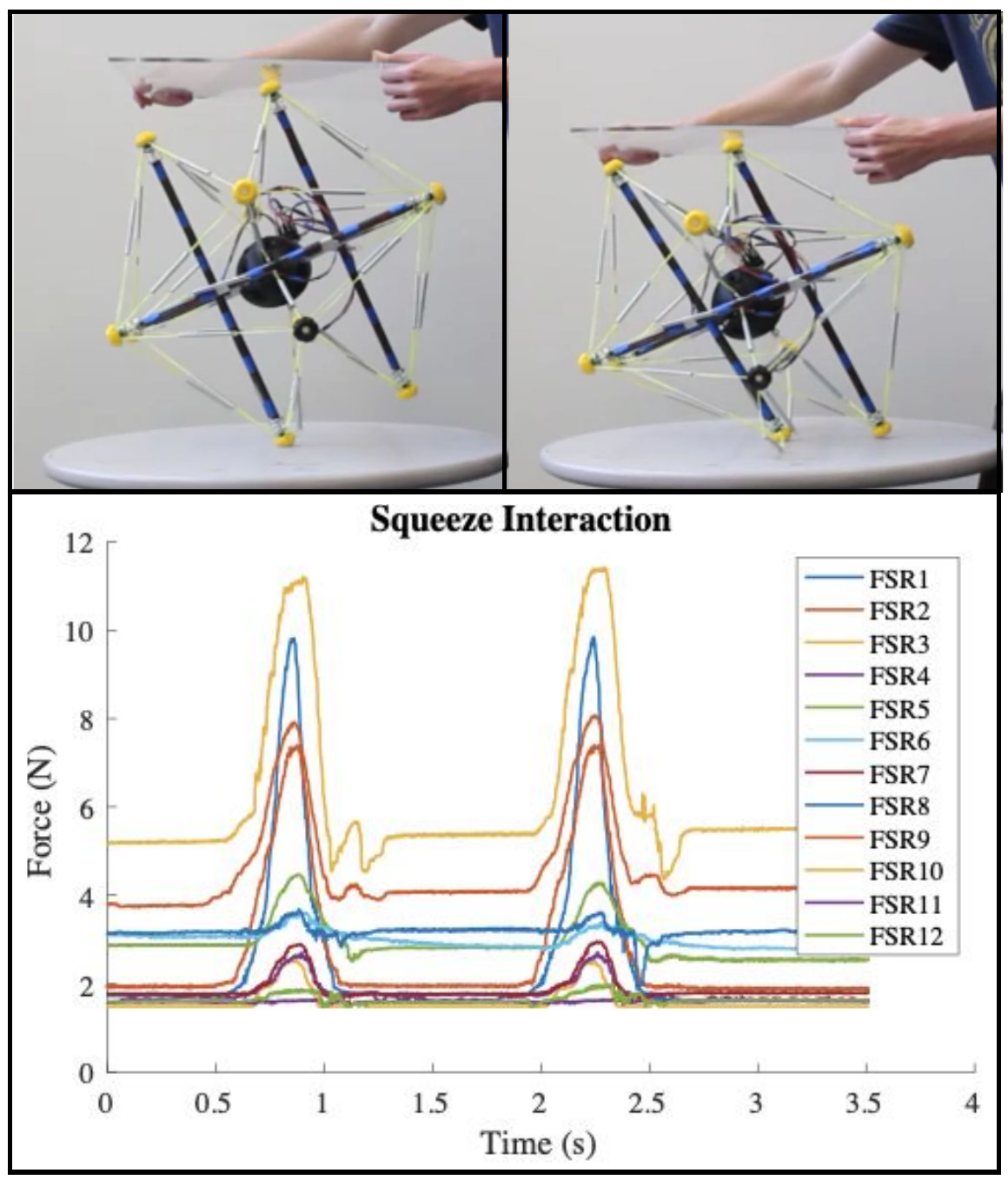}
    \caption{(top) Stills of a squeeze interaction experiment with the force-sensing tensegrity. (bottom) Force sensor signals from a squeeze interaction experiment.}
    \label{fig:exp}
\end{figure}

In all interactions, we assign a random starting orientation of the tensegrity in order to provide our classification framework with a variety of potential signal profiles from the array of force sensors. Timestamped force data for all 12 sensors is recorded on a microSD card at $60$Hz by a custom-built embedded instrumentation board. Data is also transmitted wirelessly at approximately $10$Hz to a control PC that monitors sensor performance and experiment logistics. Following sequences of experiments, data sets are manually truncated based on interaction timings, labelled according to interaction type, and sectioned into observations using a set time window size. Figure \ref{fig:exp} shows an example {\it squeeze} interaction and a recorded set of force sensor signals.

\section{Results}

As discussed in our classification modeling framework, we will be examining several combinations of processing approaches and algorithms in order to fully assess the feasibility of our pHRI sensing approach with the prototype tensegrity. In the processing stage, we section observations by window sizes in intervals of 10 samples up to 100 samples, resulting in varying counts of total observation data sets. Observation counts for each class for a representative window size of 60 samples (1 sec.) are shown in Table \ref{samples}.

\begin{table}[ht]
\centering
\caption{Observation counts by class of interaction for window size of 60 samples.}
\label{samples}
\renewcommand{\arraystretch}{1.5}
\begin{tabular}{|>{\centering\bfseries}m{2cm}|>{\centering\arraybackslash}m{1cm}|>{\centering\arraybackslash}m{1cm}|>{\centering\arraybackslash}m{1cm}|>{\centering\arraybackslash}m{1cm}|}
\hline
Class & {\bf Null} & {\bf Drop} & {\bf Squeeze} & {\bf Handle}\\
\hline
\addstackgap{Counts} & \addstackgap{3930} & \addstackgap{2643} & \addstackgap{4648} & \addstackgap{539}\\
\hline
\end{tabular}
\end{table}

In order to mitigate effects from class imbalance, we apply the synthetic minority oversampling technique (SMOTE) iteratively to reach an even distribution of samples across classes \cite{chaw2002}. The algorithms are trained using repeated stratified k-fold cross validation, and resulting performance metrics are averaged across folds. Metrics including precision, recall, and F1 scores within classes for the most well-performing window size for each classification algorithm are presented in Table \ref{results}.

\begin{table}[ht]
\centering
\caption{Classification performance metrics within classes for best-performing window size for each algorithm.}
\label{results}
\renewcommand{\arraystretch}{1.5}
\begin{tabular}{|>{\centering\bfseries}m{1.1cm}|>{\centering\arraybackslash}m{1.3cm}|>{\centering\arraybackslash}m{1.3cm}|>{\centering\arraybackslash}m{1.3cm}|>{\centering\arraybackslash}m{1.3cm}|}
\hline
\multirow{2}{*}{\Large{\sfrac{Raw}{Abst.}}} & \multicolumn{4}{c|}{{\bf k-Nearest Neighbors (KNN), Window = 100 Samples}} \\ [1ex]
\cline{2-5}
 & {\bf Null} & {\bf Drop} & {\bf Squeeze} & {\bf Handle}\\
\hline
Prec. & \fracentry{0.61}{0.78} & \fracentry{0.74}{0.65} & \fracentry{0.95}{0.90} & \fracentry{0.98}{0.68}\\
\hline
Recall & \fracentry{0.83}{0.88} & \fracentry{0.61}{0.79} & \fracentry{0.78}{0.66} & \fracentry{0.89}{0.89}\\
\hline
F1 & \fracentry{0.70}{0.82} & \fracentry{0.67}{0.71} & \fracentry{0.86}{0.76} & \fracentry{0.93}{0.77}\\
\hline
\end{tabular}
%
%
\begin{tabular}{|>{\centering\bfseries}m{1.1cm}|>{\centering\arraybackslash}m{1.3cm}|>{\centering\arraybackslash}m{1.3cm}|>{\centering\arraybackslash}m{1.3cm}|>{\centering\arraybackslash}m{1.3cm}|}
\hline
\multirow{2}{*}{\Large{\sfrac{Raw}{Abst.}}} & \multicolumn{4}{c|}{{\bf Random Forest (RF), Window = 10 Samples}} \\ [1ex]
\cline{2-5}
 & {\bf Null} & {\bf Drop} & {\bf Squeeze} & {\bf Handle}\\
\hline
Prec. & \fracentry{0.95}{0.97} & \fracentry{0.78}{0.97} & \fracentry{0.97}{0.99} & \fracentry{0.96}{0.96}\\
\hline
Recall & \fracentry{0.78}{0.98} & \fracentry{0.96}{0.97} & \fracentry{0.98}{0.99} & \fracentry{0.99}{0.99}\\
\hline
F1 & \fracentry{0.86}{0.97} & \fracentry{0.87}{0.97} & \fracentry{0.98}{0.99} & \fracentry{0.98}{0.98}\\
\hline
\end{tabular}
\end{table}

\begin{figure}[ht]
    \centering
    \includegraphics[width=0.48\textwidth]{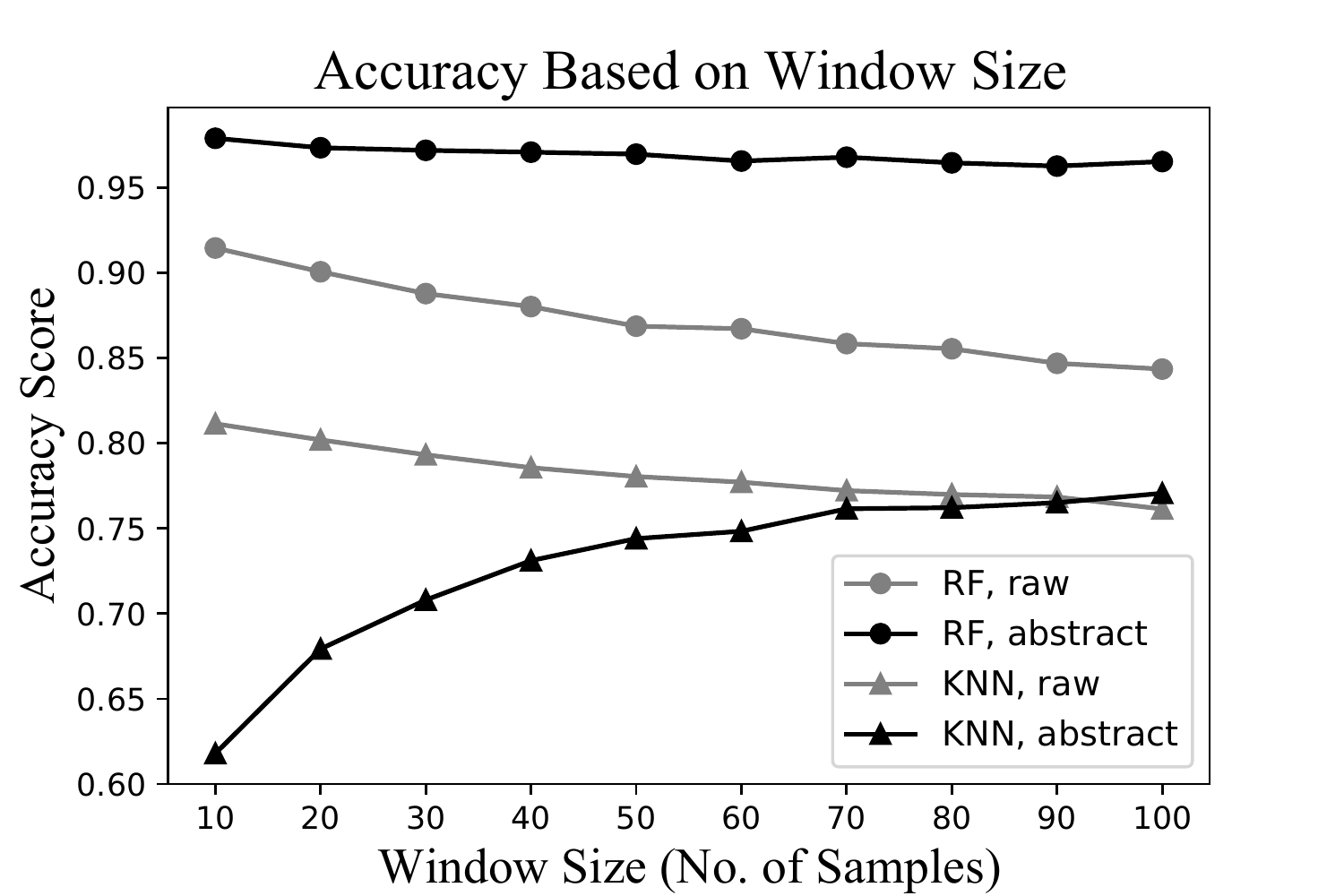}
    \caption{Accuracy comparison of algorithms and feature processing approaches across a range of window sizes.}
    \label{fig:line1}
\end{figure}

\begin{figure}[ht]
    \centering
    \includegraphics[width=0.48\textwidth]{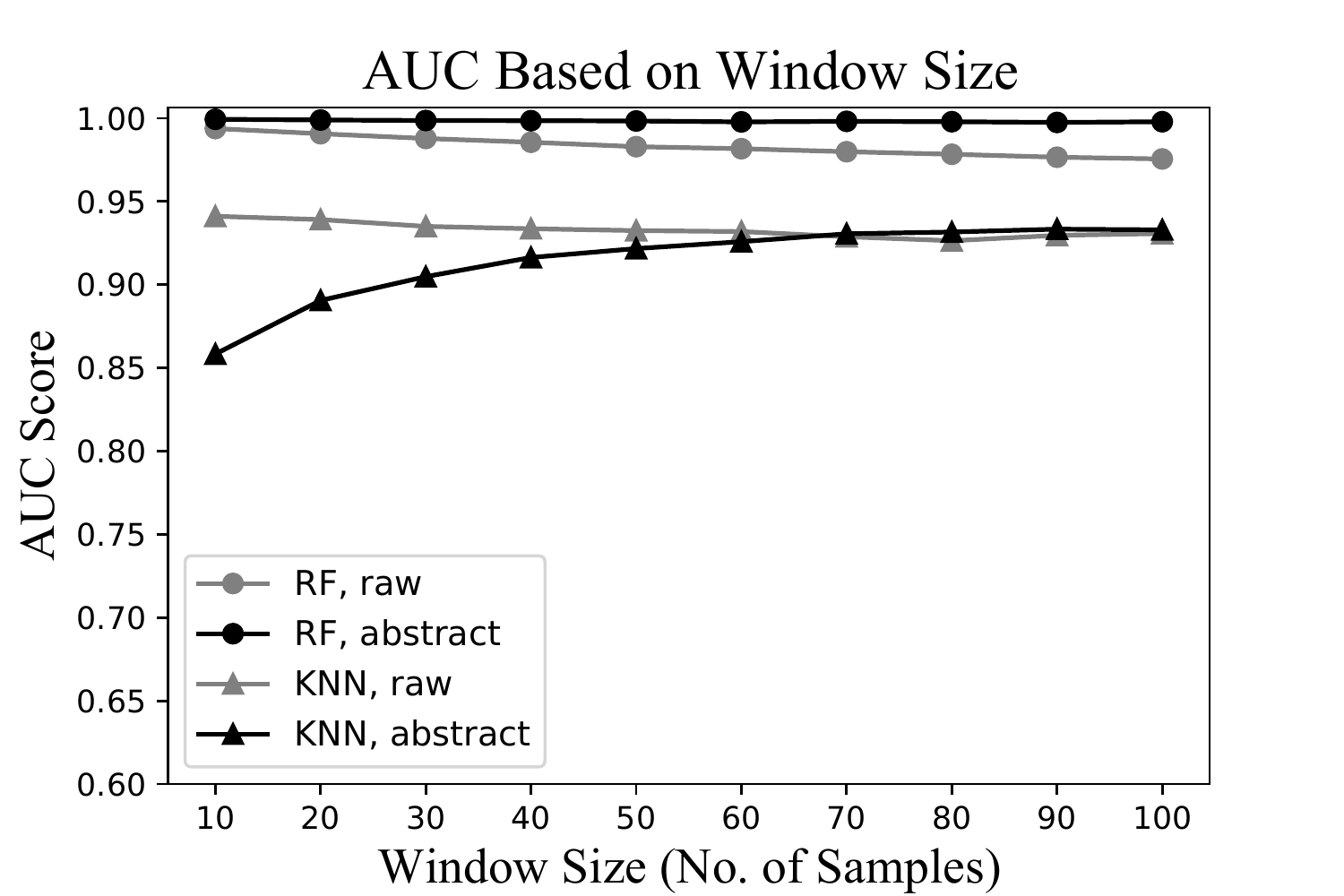}
    \caption{One-vs.-one AUC comparison of algorithms and feature processing approaches across a range of window sizes.}
    \label{fig:line2}
\end{figure}

In particular, we highlight the fact that our statistical and temporally abstracted feature engineering approach with Random Forest achieves high F1 scores across classes for the minimum window size of 10 samples. As a way to examine the effects of time dependence on the robustness of our classification approach, we compare the overall accuracy score of our classifiers to window size in Figure \ref{fig:line1}. Figure \ref{fig:line2} compares area-under-the-curve (AUC) metric, computed in a one-vs.-one approach for our multiclass case, to window size. The best performing configuration of processing and algorithm is with the ``abstract" processing approach and Random Forest algorithm at a window size of 10 samples, resulting in an accuracy of $0.98$ and an AUC score of $0.99$.

The resulting classification metrics from initial experiments clearly demonstrate the feasibility of inferring physical human-robot interaction events that are unique to contexts involving compliant robots using our novel implementation of a force-sensing tensegrity. High AUC scores across the various processing approaches and algorithms show inherent potential for class differentiation based on available information provided by force sensor data, and by extension, other feature inputs calculated from force signals as presented here. Utilization of statistically derived and temporally abstracted feature sets in the Random Forest implementation of our classification framework allows the classifier to perform robustly across varying lengths of observation windows, which indicates suitability for practical realizations of real-time inference.

\section{Conclusions and Future Work}

We presented the design of a force-sensing 6-bar spherical tensegrity that is capable of detecting a broad range of potential physical interactions with human operators using a novel integration of force-sensing resistors. By leveraging contemporary supervised learning approaches, we demonstrated the feasibility of distinguishing between a set of representative physical interaction types using a combination of processing approaches and multiclass classifier training algorithms. The results of initial experiments with said classification framework indicate high potential for reliable intent inference based on physical interaction that can be robust to temporal sensitivity.
\par
This initial investigation of physical human-robot interaction techniques with compliant robotic systems is an important step toward equipping humans and robots with more intuitive mechanisms for cooperation and collaboration through contact. In particular, we believe that the force-sensing tensegrity and subsequent iterations will be useful tools for exploring how the physical properties of compliant mobile robots can be exploited toward new types of physical interactions in a variety of human contexts. Furthermore, our experimentation with contemporary classification strategies as a framework for distinguishing interaction events instills us with confidence in the suitability of our system for increasingly complex pHRI protocols that extend to broader sets of physical interaction inputs and task use cases.
\par
Currently, we are in the process of developing a second iteration of the force-sensing tensegrity that is focused on refining integration of the force sensor array and further improving the user interface. Future experiments will be focused on exploring the intuitiveness of interaction types from the human's perspective as well as task-specific inference leveraging representative use case scenarios such as simulated disaster response in real time. We will also investigate the application of additional inference techniques including optimization-based strategies and time series classification approaches that leverage ResNets \cite{faw2019} to further improve the performance of pHRI inference with our system.

\addtolength{\textheight}{-12.75cm}   




\section*{Acknowledgment}

We would like to thank Squishy Robotics for their collaboration and consultation on the design of the force-sensing tensegrity. In addition, we want to recognize the many sponsored teams of undergraduate and graduate researchers at UC, Berkeley whose contributions were essential to the success of this project.



\begin{thebibliography}{99}

\bibitem{skel2009} R.E. Skelton, M.C. De Oliveira, {\it Tensegrity systems}, Vol. 1. New York: Springer, 2009.
\bibitem{sabel2015} A.P. Sabelhaus, J. Bruce, K. Caluwaerts, P. Manovi, R.F. Firoozi, S. Dobi, A.M. Agogino, V. Sunspiral. ``System design and locomotion of SUPERball, an untethered tensegrity robot," in {\it 2015 IEEE Intern. Conf. Robot. Automat. (ICRA)}, 2015, pp. 2867-2873.
\bibitem{fries2014} J. Friesen, A. Pogue, T. Bewley, M. de Oliveria, R. Skeltonb, V. Sunspiral. ``DuCTT: A tensegrity robot for exploring duct systems," in {\it 2014 IEEE Intern. Conf. Robot. Automat. (ICRA)}, 2014, pp. 4222-4228.
\bibitem{bruce2014} J. Bruce, K. Caluwaerts, A. Iscen, A.P. Sabelhaus, V. Sunspiral. ``Design and evolution of a modular tensegrity robot platform," in {\it 2014 IEEE Intern. Conf. Robot. Automat. (ICRA)}, 2014, pp. 3483-3489.
\bibitem{booth2020} J.W. Booth, O. Cyr-Choinière, J.C. Case, D. Shah, M.C. Yuen, R. Kramer-Bottiglio, ``Surface Actuation and Sensing of a Tensegrity Structure Using Robotic Skins," {\it Soft Robot.}, Sep. 2020.
\bibitem{calan2014} A. Calanca, L. Capisani, P. Fiorini, ``Robust force control of series elastic actuators," {\it Actuators}, vol. 3, no. 3, pp. 182-204, 2014.
\bibitem{yu2015} H. Yu, S. Huang, G. Chen, Y. Pan, Z. Guo, ``Human–Robot Interaction Control of Rehabilitation Robots With Series Elastic Actuators," in {\it IEEE Trans. Robot.}, vol. 31, no. 5, pp. 1089-1100, Oct. 2015.
\bibitem{lange2013} F. Lange, W. Bertleff, M. Suppa, ``Force and trajectory control of industrial robots in stiff contact." in {\it 2013 IEEE Intern. Conf. Robot. Automat. (ICRA)}, 2013, pp. 2927-2934.
\bibitem{argall2010} B.D. Argall, A.G. Billard, ``A survey of tactile human–robot interactions," {\it Robot. Auton. Sys.}, vol. 58, no. 10, pp. 1159-1176, Oct. 2010.
\bibitem{ciri2015} A. Cirillo, F. Ficuciello, C. Natale, S. Pirozzi, L. Villani, ``A conformable force/tactile skin for physical human–robot interaction," {\it IEEE Robot. Autom. Letters}, vol. 1, no. 1, pp. 41-48, Dec. 2015.
\bibitem{sant2008} A. De Santis, B. Siciliano, A. De Luca, A. Bicchi, ``An atlas of physical human–robot interaction," {\it Mech. Mach. Theory}, vol. 43, no. 3, pp. 253-270, Mar. 2008.
\bibitem{hadd2016} S. Haddadin, E. Croft, {\it Physical human-robot interaction}. Cham: Springer, 2016.
\bibitem{losey2018} D.P. Losey, C.G. McDonald, E. Battaglia, M.K. O'Malley, ``A review of intent detection, arbitration, and communication aspects of shared control for physical human–robot interaction," {\it Appl. Mech. Rev.}, vol. 70, no. 1, Jan. 2018.
\bibitem{agra2014} D.J. Agravante, A. Cherubini, A. Bussy, P. Gergondet, A. Kheddar, ``Collaborative human-humanoid carrying using vision and haptic sensing," {\it 2014 IEEE Intern. Conf. Robot. Automat. (ICRA)}, 2014, pp. 607-612.
\bibitem{bajc2018} A. Bajcsy, D.P. Losey, M.L. O'Malley, A.D. Dragan, ``Learning from physical human corrections, one feature at a time," in {\it Proc. 2018 ACM/IEEE Intern. Conf. Hum.-Robot Inter.}, 2018, pp. 141-149.
\bibitem{sanan2011} S. Sanan, M.H. Ornstein, C.G. Atkeson, ``Physical human interaction for an inflatable manipulator," in {\it 2011 Ann. Intern. Conf. IEEE Engin. Med. Bio. Soc.}, 2011, pp. 7401-7404.
\bibitem{poly2017} P. Polygerinos, N. Correll, S.A. Morin, B. Mosadegh, C.D. Onal, K. Petersen, M. Cianchetti, M.T. Tolley, R.F. Shepherd, ``Soft robotics: Review of fluid‐driven intrinsically soft devices; manufacturing, sensing, control, and applications in human‐robot interaction," {\it Adv. Eng. Mat.}, vol. 19, no. 12, Dec. 2017.
\bibitem{maj2014} C. Majidi, ``Soft robotics: a perspective—current trends and prospects for the future," {\it Soft Rob.}, vol. 1, no. 1, pp. 5-11, Mar. 2014.
\bibitem{squish2019} The Robot Report. (2019, May 13). {\it Squishy Robotics releases mobile robots that can be dropped from aircraft in emergencies} [Online]. Available: https://www.therobotreport.com/squishy-robotics-mobile-sensors-aircraft-drop-emergencies
\bibitem{sult2004} C. Sultan, R. Skelton, ``A force and torque tensegrity sensor," {\it Sens. Actuat. A: Phys.}, vol. 112, no. 2-3, pp. 220-231, May 2004.
\bibitem{sabel2020} A.P. Sabelhaus, A.H. Li, K.A. Sover, J.R. Madden, A.R. Barkan, A.K. Agogino, A.M. Agogino, ``Inverse Statics Optimization for Compound Tensegrity Robots," {IEEE Robot. Automat. Letters}, vol. 5, no. 3, pp. 3982-3989, Mar. 2020.
\bibitem{tur2009} J.M.M. Tur, S.H. Juan, ``Tensegrity frameworks: Dynamic analysis review and open problems," {\it Mech. Mach. Theory}, vol. 44, no. 1, pp. 1-18, Jan. 2009.
\bibitem{aly2005} M. Aly, ``Survey on multiclass classification methods," {\it Neur. Netw.}, vol. 19, pp. 1-9, Nov. 2005.
\bibitem{chaw2002} N.V. Chawla, K.W. Bowyer, L.O. Hall, W.P. Kegelmeyer, ``SMOTE: synthetic minority over-sampling technique," {\it Journal artif. intell. research}, vol. 16, pp. 321-357. Jun. 2002.
\bibitem{liaw2002} A. Liaw, M. Wiener, ``Classification and regression by randomForest," {\it R news}, vol. 2, no. 3, pp. 18-22, Dec. 2002.
\bibitem{faw2019} H.I. Fawaz, G. Forestier, J. Weber, L. Idoumghar, P.A. Muller, ``Deep learning for time series classification: a review," {\it Data Mining Knowl. Disc.}, vol. 33, no. 4, pp. 917-963, Jul. 2019.






\end{thebibliography}
\end{document}